\documentclass[letterpaper]{article} 
\usepackage{aaai17}  
\usepackage{times}  
\usepackage{helvet}  
\usepackage{courier}  
\usepackage{url}  
\usepackage{graphicx}  
\usepackage[usenames,dvipsnames]{xcolor}
\begin{document}
\newcommand{\ce}[1]{\texttt{\small{#1}}}

\title{Explainable AI for Intelligence Augmentation in Multi-Domain Operations}
\author{Alun Preece \\
Crime and Security Research Institute \\
Cardiff University, UK \\
Email: PreeceAD@cardiff.ac.uk
\And 
Dave Braines \\
IBM Research \\ 
Hursley, Hampshire, UK \\
Email: dave\_braines@uk.ibm.com
\AND
Federico Cerutti \\
Crime and Security Research Institute \\
Cardiff University, UK \\
Email: CeruttiF@cardiff.ac.uk
\And 
Tien Pham \\
U.S. Army Research Laboratory \\
Adelphi, MD, USA \\
Email: tien.pham1.civ@mail.mil
}

\newcommand{\aedit}[1]{{\color{black} #1}}

\maketitle

\begin{abstract}
Central to the concept of multi-domain operations (MDO) is the utilization of an intelligence, surveillance, and reconnaissance (ISR) network consisting of overlapping systems of remote and autonomous sensors, and human intelligence, distributed among multiple partners. Realising this concept requires advancement in both artificial intelligence (AI) for improved distributed data analytics and intelligence augmentation (IA) for improved human-machine cognition. The contribution of this paper is threefold: (1) we map the coalition situational understanding (CSU) concept to MDO ISR requirements, paying particular attention to the need for assured and explainable AI to allow robust human-machine decision-making where assets are distributed among multiple partners; (2) we present illustrative vignettes for AI and IA in MDO ISR, including human-machine teaming, dense urban terrain analysis, and enhanced asset interoperability; (3) we appraise the state-of-the-art in explainable AI in relation to the vignettes with a focus on human-machine collaboration to achieve more rapid and agile coalition decision-making. The union of these three elements is intended to show the potential value of a CSU approach in the context of MDO ISR, grounded in three distinct use cases, highlighting how the need for explainability in the multi-partner coalition setting is key.
\end{abstract}

\section{Introduction}

Multi-domain operations (MDO) require the capacity, capability, and endurance to operate across multiple domains --- from dense urban terrain to space and cyberspace --- in contested environments against near-peer adversaries~\cite{TRADOC:2018}. A key characteristic of the operational environment in MDO is that adversaries will be contesting all domains, the electromagnetic spectrum, and the information environment, and allied dominance is not assured. \aedit{Adversaries attempt to achieve \emph{stand-off} by separating friendly forces in multiple dimensions: temporally, spatially, functionally, and politically. Stand-off is achieved by reducing allies' speed of recognition, decision and action, as well as by fracturing alliances through multiple means: diplomatic, economic, conventional and unconventional warfare, including information warfare.} In this context, rapid and continuous integration of capabilities to collect, process, disseminate and exploit actionable information and intelligence becomes more critical than ever before. 

Addressing this challenge, the concept of \textit{layered ISR} in MDO envisions exploitation of `an existing intelligence, surveillance, and reconnaissance (ISR) network developed with partners\ldots that consists of overlapping systems of remote and autonomous sensors, human intelligence, and friendly special operations forces' (\cite{TRADOC:2018}, pp.33--34). Maximising the value of ISR assets in the unprecedentedly-contested environment requires an ability to share resources among partners --- in operations conducted as part of joint, interagency, and multinational teams --- in a controlled but open coalition environment, with knowable levels of trust and confidence. 

Artificial intelligence (AI) and machine learning (ML) techniques are seen as key to realising the layered ISR vision in MDO: `rapidly disseminating data to a field army or corps analysis cell employing artificial intelligence or other computer assistive technologies to analyze the high volume of data' (\cite{TRADOC:2018}, pp.39). Indeed, the demands of the MDO environment are seen as requiring an ability to converge capabilities --- including ISR --- spanning multiple domains at a speed and scale that exceeds human cognitive abilities. Robust and interoperable AI/ML is viewed as key in fusing data from multiple assets and disseminating actionable knowledge across operation partners to inform decision-making and task accomplishment~\cite{Spencer:2019}.

In summary, the challenge is to enable humans and machine agents (software and robotic) to operate effectively in joint, interagency, multinational and highly-dispersed teams, in distributed, dynamic, complex, and cluttered environments. From the humans' perspective, AI and ML are necessary tools to overcome human cognitive limits due to the speed and scale of operations, with the goal of \textit{augmenting} --- not replacing --- human cognition and decision-making. Here we view \textit{intelligence augmentation} (IA) as a complement to AI, as first envisioned in the earliest period of AI's history~\cite{Engelbart:1962}. We focus on rapidly-formed coalition teams comprised of human and AI/ML agents, operating at the edge of a network, with limited connectivity, bandwidth and compute resources, in a decision-making role, for example, by Army Soldiers in a dense urban setting. However, much of the discussion will also apply to a range of other roles in other domains, for example, intelligence analysts conducting cyber-domain decision-making.

We have previously studied this challenge in a related context: that of \textit{coalition situational understanding} (CSU)~\cite{Preece:2017} wherein we identified two particular properties of importance in human-machine collaboration: \textit{explainability} to underpin confidence and \textit{tellability} to improve operational agility and performance. This paper focuses mainly on the first of these, but also touches upon the second. We begin by revisiting the CSU concept in the MDO context, before examining how the concept applies in the context of three MDO vignettes: human-machine teaming, dense urban terrain analysis, and enhanced asset interoperability. Finally, we appraise the state-of-the-art in explainable AI in relation to the vignettes, highlighting how the notion of \textit{layered explanations}~\cite{Preece:2018} is well-fitted to the need for AI/ML assurance in MDO layered ISR.

Before proceeding, we step back and note that key features of MDO environments --- (i) rapidly changing situations; (ii) limited access to real data to train AI; (iii) noisy, incomplete, uncertain, and erroneous data inputs during operations; and (iv) peer adversaries that employ deceptive techniques to defeat algorithms --- are not unique to the military context; they are often found in government and public sector applications more generally, as are the joint, interagency, and multinational aspects of these endeavours. Indeed, in general, the multi-domain breadth of the MDO concept and its consideration of both competition and conflict phases, means that MDO impinges on the political and societal spheres that are the province of government and the public sector.

\section{Coalition Situational Understanding for MDO}

Situational understanding (SU) is the `product of applying analysis and judgment to the unit's situation awareness to determine the relationships of the factors present and form logical conclusions concerning threats to the force or mission accomplishment, opportunities for mission accomplishment, and gaps in information'~\cite{Dostal:2007}. UK military doctrine~\cite{JDP04:2010} defines \textit{understanding} in the following terms:

{\small 
\noindent \textbf{Comprehension} (insight) = situational awareness and analysis \\
\textbf{Understanding} (foresight) = comprehension and judgement}

Here, understanding includes \textit{foresight}, i.e., an ability to infer (predict) potential future states, which is compatible with the common definition that SU involves being able to draw conclusions concerning threats~\cite{Dostal:2007}. Foresight necessarily includes an ability to process and reason about information temporally. These views of SU are intrinsically linked to information fusion in that they involve the collection and processing of data from multiple environmental sources as input to deriving SU. In terms of the \aedit{JDL (Joint Directors of Laboratories) Model of data fusion}~\cite{Blasch:2006}, a CSU problem may address relatively high or relatively low levels of understanding, in terms of the kinds of semantic entities and relationships considered. For example, at the relatively low levels a CSU problem may be concerned with only the detection, identification and localization of objects such as vehicles or buildings (JDL Levels~1 and~2). At higher levels, a CSU problem would be concerned with determining threats, intent, or anomalies (JDL Level~3). Moreover, the sources will commonly span multiple modalities, for example, imagery, acoustic and natural language data~\cite{Lahat:2015}.

\begin{figure}[t]
\centering
\includegraphics[width=0.4\textwidth]{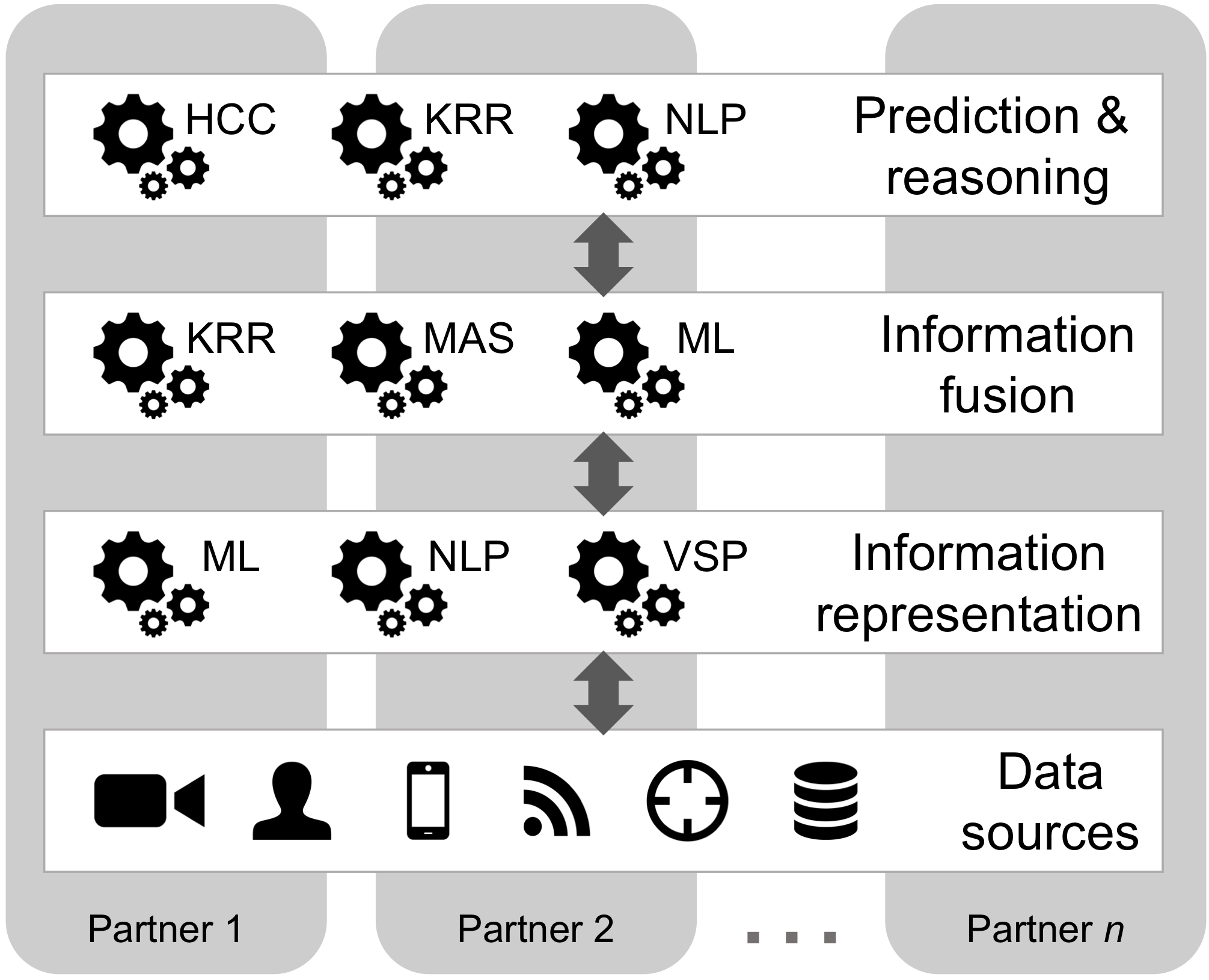}
\caption{CSU layered model (from~\cite{Preece:2017}) distributed virtually across multiple partners and employing multiple technologies: human-computer collaboration (HCC), knowledge representation and reasoning (KRR); multi-agent systems (MAS);  machine learning (ML); natural language processing (NLP), vision and signal processing (VSP).}
\label{fig:layers}
\end{figure}

Our conceptual architecture for SU in a coalition operations context --- coalition situational understanding (CSU) --- is illustrated in Figure~\ref{fig:layers}. The lowest layer consists of a collection of data sources (physical sensors and human-generated content), accessible across the coalition, collecting multimodal data. The three upper layers roughly correspond to Levels~0--3 of the JDL Model. For each layer, the figure shows the primary technologies --- including AI and ML --- employed, though others may be exploited also. The information representation layer uses incoming data streams to learn concepts and model entities together with their relationships at multiple levels of semantic granularity. The history of past observations is encoded in these representations, explicitly or implicitly. The information fusion layer employs algorithms and techniques developed to perform fusion over concepts and entities derived from the information representation layer. This layer estimates the current state of the world, providing \textit{insight} (situational awareness). The prediction and reasoning layer then uses the estimated current state, together with the state space of the models to predict the future state, providing \textit{foresight} (situational understanding). The figure depicts a virtual view of the coalition: all four layers are distributed across the coalition. 

In accordance with the User Fusion model~\cite{Blasch:2006}, the upper layers in Figure~\ref{fig:layers} need to be open to humans to provide expert knowledge for reasoning; these layers also need to be open to the human user in terms of being able to generate explanations of the insight and foresight generated by the system. There is a bi-directional exchange of information occurring between the different layers: in the upward (feedforward) direction, the inferences at the lower layer act as input for the next higher layer; in the downward (feedback) direction, information is used to adjust the model and algorithm parameters and possibly task the sensors differently. Creating better systems to support CSU necessitates the development of mature models and algorithms that can over a period of time reduce the human intervention and attain greater autonomy, but without replacing human involvement and oversight.

\section{CSU-MDO Vignettes}

In this section, we consider three vignettes in the context of CSU in MDO. The first examines the need for robust and interoperable AI/ML services; the second examines the dynamics of human-machine collaboration, and the third considers the focus on dense urban terrain operations.

\subsection{Vignette 1: Enhanced Asset Interoperability}

\begin{figure}[t]
\centering
\includegraphics[width=0.4\textwidth]{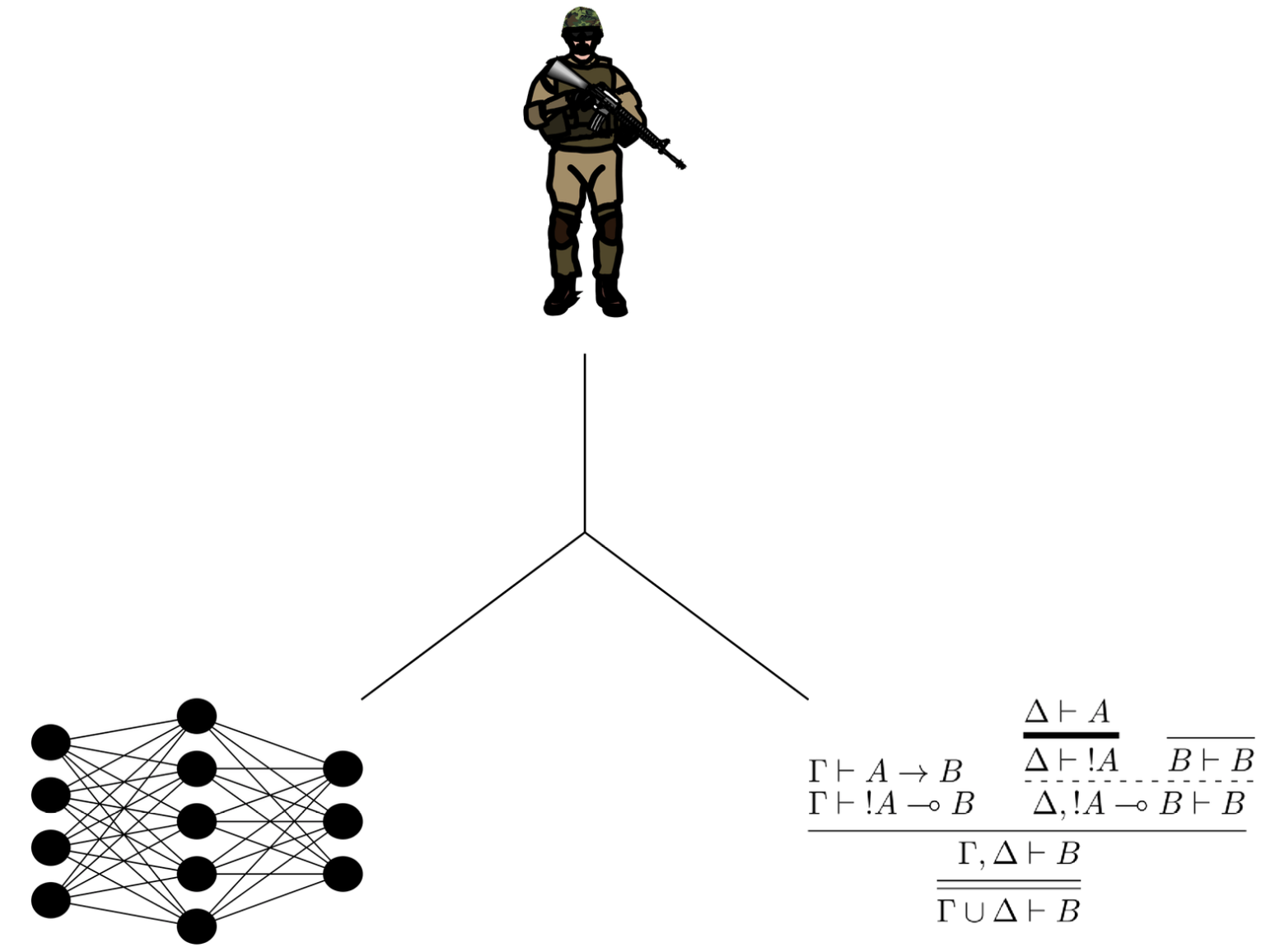}
\caption{A multi-agent non-hierarchical approach to CSU: (top) human agent, (bottom left) subsymbolic AI agent, (bottom right) symbolic AI agent.}
\label{fig:nonhierarchical}
\end{figure}

Taking the MDO concept of layered ISR as a starting point (`overlapping systems of remote and autonomous sensors, human intelligence, and friendly special operations forces'~\cite{TRADOC:2018} p.34), we take the view that humans are one of three kinds of ISR agent in a multi-agent setting as depicted in Figure~\ref{fig:nonhierarchical}, along with software agents based on (i) subsymbolic AI technologies (e.g., deep neural networks~\cite{LeCun:2015}) and (ii) symbolic AI technologies (e.g., logic-based approaches). To achieve interoperability between these three kinds of agent (ISR asset), we need to:
\begin{enumerate}
\item enable subsymbolic AI agents to share uncertainty-aware representations of insights and knowledge that can then be communicated to symbolic AI agents;
\item equip symbolic AI agents to learn the uncertainty distribution of causal links from data, while being able to share insights to subsymbolic AI agents; and
\item develop symbiotic AI techniques to effectively interact with humans, at first by adapting stereotypical behaviours via continuous learning from human-machine teaming activities.
\end{enumerate}

The first two cases focus on interoperability between machine assets. In the third case we go beyond the traditional hierarchical architecture that sees humans interacting only with symbolic AI-equipped agents that in turn leverage subsymbolic AI for achieving human-level or superior performance on specific tasks. Such a traditional architecture is limited because: (1) it is not always the case that symbolic AI is needed for interaction with humans~\cite{Ribeiro:2016}; (2) there are tasks for which a symbolic AI can support a subsymbolic AI agent~\cite{Xu:2018}; and (3) there are tasks for which humans can support symbolic and/or subsymbolic AI agents~\cite{Phan:2016}, hence AI agents need to be equipped with the capabilities to learn and reason about human hierarchies and structures.

\begin{figure}[th]
\centering
\includegraphics[width=0.46\textwidth]{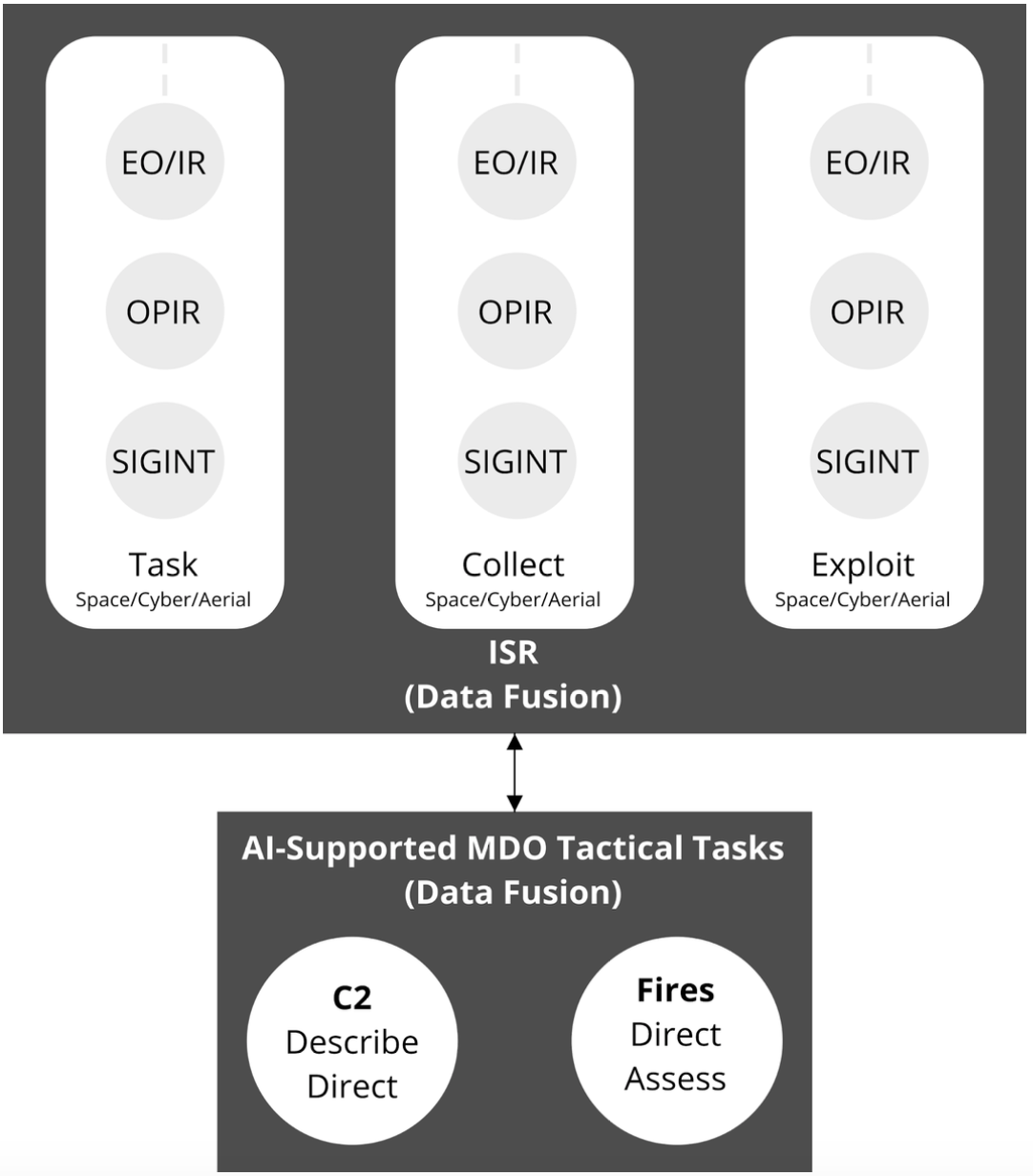}
\caption{Simplified version of the figure from~\cite{Spencer:2019}: rectangles represent symbolic systems; circles represent subsymbolic systems; rounded rectangles represent hybrid elements.}
\label{fig:mdo}
\end{figure}

Figure~\ref{fig:mdo} provides a mapping between the MDO layered ISR architecture envisioned in~\cite{Spencer:2019} and the preceding characterisation of assets as symbolic, subsymbolic, or hybrid.

\subsection{Vignette 2: Human-Machine Teaming}

Our work seeks to advance capabilities to contribute to complex coalition tasks in support of MDO, where the need for joint and multinational teams and multi-domains is cardinal~\cite{TRADOC:2018}. It is of paramount importance to provide a coherent view and assessment of operational situations as they happen thus integrating learning and reasoning for CSU in complex, contested environments to inform decision makers at the edge of the network. As previously noted, CSU requires both collective insight --- accurate and deep understanding of a situation derived from uncertain and often sparse data, and collective foresight --- the ability to predict what will happen in the future~\cite{Preece:2017}. 

The notion of \textit{affordances} has been central to the human-computer interaction (HCI) field for many years, referring to what an object `is for', i.e., `the perceived and actual properties of the thing, primarily those fundamental properties that determine just how the thing could possibly be used'\cite{Norman:1988}. In the MDO layered ISR context, it is necessary to consider human and machine assets in terms of their affordances to a range of ISR tasks. The purpose of human-machine teaming is to aim for each party to exploit the strengths of, and compensate for the weaknesses of, the other~\cite{Cummings:2014}. For example,~\cite{Crouser:2012} characterised machine affordances in the scope of visual analytics as follows:
\begin{itemize}
\item large-scale data manipulation;
\item collection and storage of large data volumes;
\item efficient data movement;
\item bias-free analysis.
\end{itemize}

\noindent 
Based on current machine capabilities, the following constitute human asset affordances~\cite{Crouser:2012}:
\begin{itemize}
\item visual and audiolinguistic perception;
\item sociocultural awareness;
\item creativity;
\item broad background domain knowledge.
\end{itemize}

\begin{figure}[th]
\centering
\includegraphics[width=0.47\textwidth]{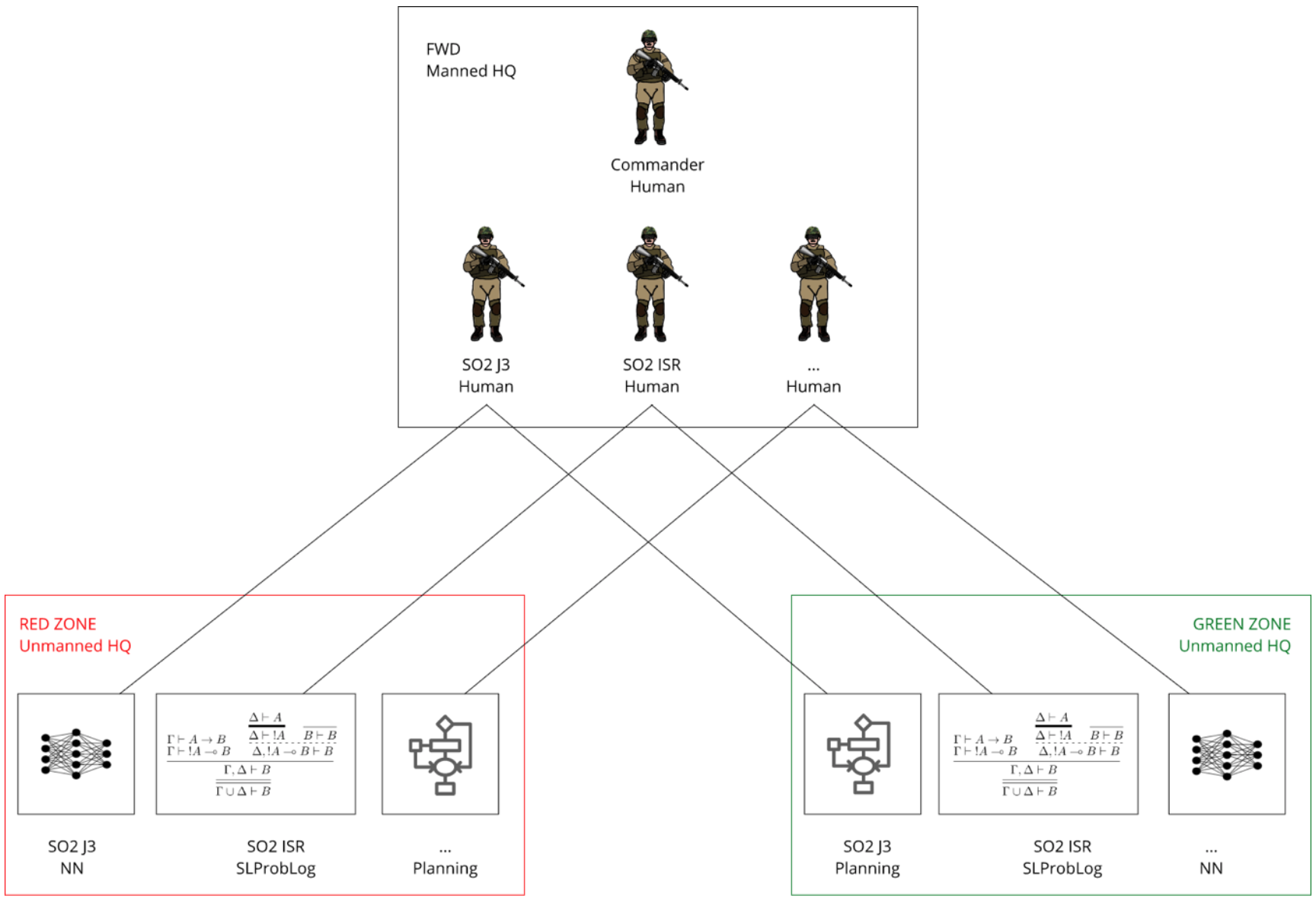}
\caption{Human-machine teaming in the tactical domain: deployment of  manned and unmanned tactical headquarters equipped with subsymbolic and symbolic AI agents; elaborated from~\cite{White:2019}.}
\label{fig:hmt}
\end{figure}

In fulfilling the MDO, it will become common to envisage deployment of both manned and unmanned tactical headquarters (HQ) as illustrated in Figure~\ref{fig:hmt}, elaborated from the scenario in~\cite{White:2019}. Here, concurrently with the deployment of the manned HQ~A, a second unmanned HQ~B is established further forward in a high threat area consisting of `virtual staff officers'. These are designed to work in cohort with their opposite numbers in the manned HQ and reduce both HQ footprint as well as the workload of --- and threat to --- human operators. A mix of both autonomous and manned sensors feed into the unmanned HQ and human-machine teaming provides the enduring requirement to have a `human in the loop' in order to make key final decisions.

\subsection{Vignette 3: Dense Urban Terrain Analysis}

Accelerating rates of urbanisation globally, and the strategic importance of cities and megacities, ensures that MDO operations will take place within dense urban terrain. Here, density refers to both the physical and demographic nature of this environment, giving rise to specific physical, cognitive, and operational characteristics. Preparation for MDO in dense urban terrain necessitates intelligence activities to understand human, social, and infrastructure details; such areas are characterised by diverse, interconnected human and physical networks, and three-dimensional engagement areas affording varying levels of ready-made cover and concealment.

In such environments, ISR will exploit and augment civilian infrastructure. For example, use of civilian CCTV (closed circuit television cameras) will increasingly be augmented with processing for automatic facial recognition for the detection and tracking of high-value targets, or to support building patterns of life. As targets move into vehicles, civilian automatic number plate recognition technology may be exploited. The diversity of such urban infrastructure --- in some cases extending to full-scale `smart city' integration --- establishes further requirements for agile interoperability between ISR assets, especially since ISR tasks cannot necessarily plan in advance what collection and processing will be needed. In such cases, analytics composition will be dynamic and context-specific, with continual re-provisioning and resource optimisation~\cite{White:2019}.

In dense urban terrain, the need for joint, interagency, and often multinational cooperation is further pronounced. As above, CSU in this context depends on human-AI collaboration: machine processes such as AI agents offer powerful affordances in terms of data analytics, but they need to provide levels of assurance (explanation, accountability, transparency) for their outputs, particularly where those outputs are consumed by decision makers without technical training in information science, and who may be exploiting relatively unfamiliar local ISR assets. Current ML approaches are limited in their ability to generate interpretable models (i.e., representations) of the world necessary for CSU~\cite{Lake:2017}. Moreover, these approaches require large volumes of training data and lack the ability to learn from small numbers of examples as people and knowledge representation-based systems do~\cite{Guha:2015}. An ability for human experts to tell a machine relevant information --- often from their lived experience of the local environment --- increases the tempo and granularity of human-AI interactions and the overall responsiveness of the system in meeting mission requirements. It is therefore important to equip coalition machine agents with integrated learning and knowledge representation mechanisms that support CSU while affording assurance (explainability) and an ability to be told key information to mitigate issues with sparse data (tellability). In our recent research we have built significant foundations for the neuro-symbolic hybrid environment, including multi-agent learning with multimodal data~\cite{Xing:2018}, evidential deep learning~\cite{Sensoy:2018}, probabilistic logic programming~\cite{Cerutti:2019}, forward inferencing architectures where the output of a neural network was fed into probabilistic logic engine to detect events with complex spatiotemporal properties~\cite{Vilamala:2019}.

\section{Layered Explanations for Layered ISR in MDO}

The emergent goal from the three vignettes in the previous section is to address the challenge of enabling rapid exploitation of adaptive ISR knowledge to inform decision-making across coalitions in MDO, by creating system architectures to enable synergistic collaboration between machine and human agents for actionable insight and foresight in a contested environment.

\begin{figure}[th]
\centering
\includegraphics[width=0.4\textwidth]{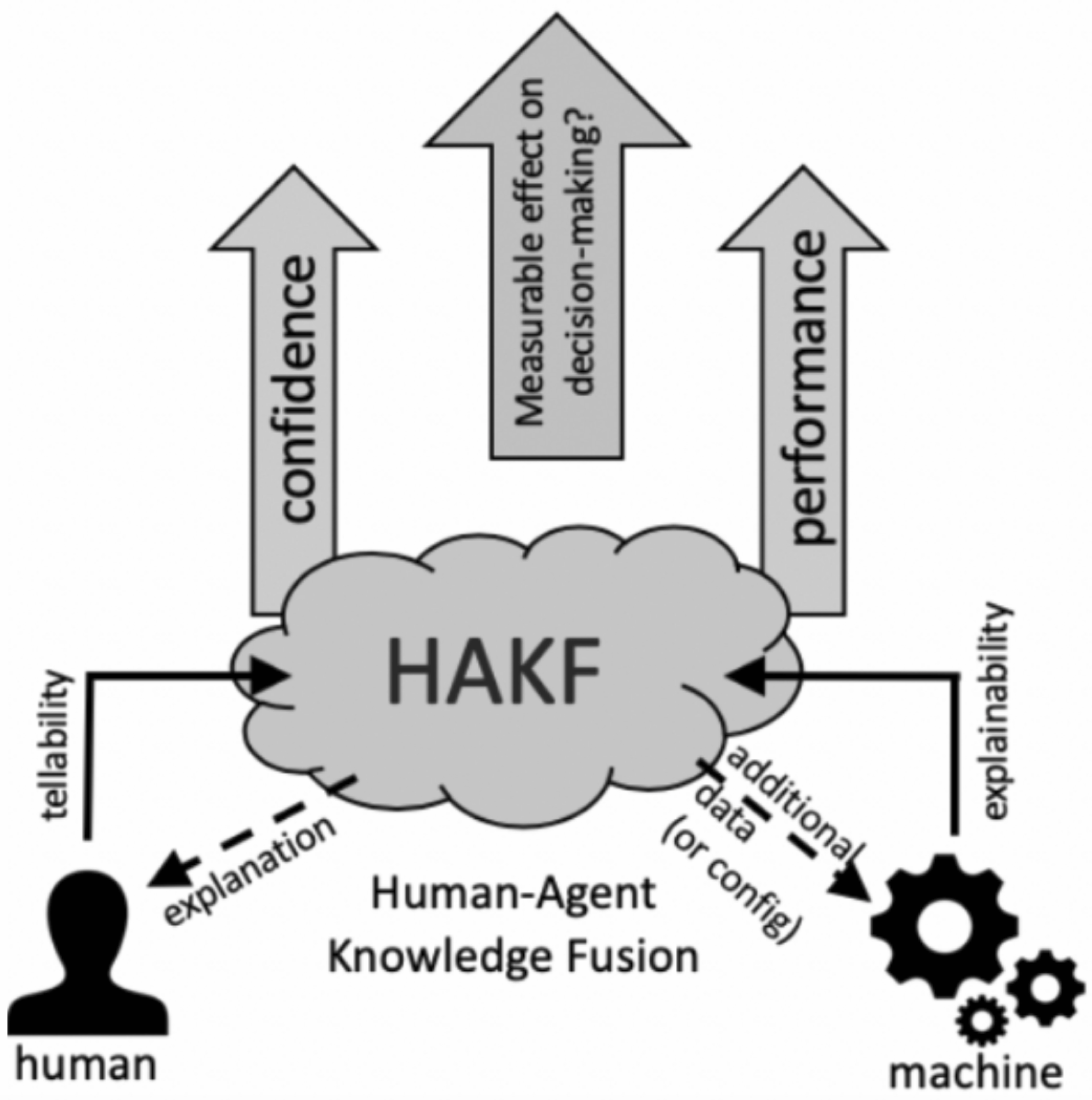}
\caption{Human-agent knowledge fusion for improved confidence and performance in support of better decision-making.}
\label{fig:hakf}
\end{figure}

Throughout our earlier research into CSU we have identified the need for the agile integration of human and machine agents from across coalition partners into dynamic and responsive teams. We have We have formalised this as  \textit{human-agent knowledge fusion} (HAKF): a capability to support this deep interaction, comprising bi-directional information flows of explainability and tellability thereby enabling meaningful communication between AI and humans~\cite{Braines:2018} as shown in Figure~\ref{fig:hakf}. This HAKF capability supports explainability and tellability naturally as conversational processes between human and machine agents~\cite{Tomsett:2018}, enabling AI agents to provide explanations of results arising from complex machine/deep learning classifications, and to receive knowledge that modifies their models or knowledge bases.

A key requirement is to add human interaction to the distributed symbolic/subsymbolic integration highlighted in the previous section and establish the minimum set of common language that the various human and AI agents need to master to ensure effective communication for a given task. To support intuitive machine processable representations in the context of dynamic context-aware gathering and information processing services, we pay particular attention to the human consumability of machine generated information, especially in the context of conversational interaction and where decision makers may lack deep technical training in information science. This common language must be capable of conveying uncertainty and the appropriate structures to achieve integration with the subsymbolic layers, as well as more traditional semantic features relevant to the domain. We do not limit ourselves to purely linguistic forms; novel visual or diagrammatic notations, or indeed other communication techniques, may be relevant as part of the solution.

Moreover, it is necessary to consider the case of automated negotiations between various autonomous agents, some of which will be humans. At the same time, humans themselves can be the object of a learning task: their own behaviour can potentially be nudged in specific directions if the machine agent learns enough about the individual human agent (or human agents in general) to infer the impact of suggestions or changes. In addition, machine agents might need to identify the best fit among the human agents for a given task, with historical data helping them towards this goal. Such symbiotic AI techniques can be used to more effectively interact with the humans, at first by adapting stereotypical behaviours via continuous learning from human-machine interactions.

Such a complex and dynamic hybrid setting is particularly risky and prone to exploitation in a contested environment, hence the need to integrate the uncertainty-aware and probabilistic capabilities. All of this much be achieved in a tempo that is appropriate to the decision-making task and the involvement of the human users, with machine agents able to support real-time interaction.

\subsection{Layered Explanations}

In recent work, we examined explainability from the perspective of explanation recipients, of six kinds~\cite{Tomsett:2018}: system \textit{creators}, system \textit{operators}, \textit{executors} making a decision on the basis of system outputs, \textit{decision subjects} affected by an executor's decision, \textit{data subjects} whose personal data is used to train a system, and system \textit{examiners}, e.g., auditors or ombudsmen. Based on this framework, we proposed a `layered' approach to offer different explanations tailored to the different stakeholders~\cite{Preece:2018} by means of a composite \textit{explanation object} that packs together all the information needed to satisfy multiple stakeholders, and can be unpacked (e.g., by accessor methods) per a recipient's particular requirements. We view such an object as being layered as follows:\\
\textbf{Layer 1 --- traceability}: transparency-based bindings to internal states of the model so the explanation isn't entirely a post-hoc rationalisation and shows that the system `did the thing \aedit{\emph{right}}'; \\
\textbf{Layer 2 --- justification}: post-hoc representations (potentially of multiple modalities) linked to layer~1, offering semantic relationships between input and output features to show that the system `did the \aedit{\emph{right}} thing'; \\
\textbf{Layer 3 --- assurance}: post-hoc representations (again, potentially of multiple modalities) linked to layer~2, with explicit reference to policy/ontology elements required to give recipients confidence that the system `does the right thing'.

\subsection{Integrated Example}

We consider a dense urban terrain setting drawing on~\cite{Kaplan:2018} in which civilian sensing infrastructure, including CCTV, is supplemented by coalition ISR assets. Video feeds from a public marketplace are being monitored using activity recognition AI/ML services as elaborated in~\cite{Vilamala:2019}. An outbreak of anomalous, `violent' physical activity is suddenly detected in the CCTV feed. At this point, via enhanced asset interoperability, the coalition ISR system accesses on demand other sensing modalities to obtain more data on the situation, tapping into recently-collected audio data from the marketplace, obtained via acoustic sensors. Processing the relevant part of the audio stream reveals rhythmic chanting that, fused together with the visual activity, signifies that the activity is a harmless dance ritual specific to the region. Note that the inference that the activity is non-threatening constitutes situational understanding: insight with foresight. Moreover, while it is conceivable that, when enough data is available to classify the activity, the harmless dance could be identified by machine processing, in~\cite{Kaplan:2018} we consider the case where recognising this activity requires local cultural knowledge and is handled by human-machine teaming: the machine brings the anomalous visual activity, including the extra context from the audio, to the attention of an experienced human agent. 

Our concept of layered explanations supports the `packing' of three levels of explanation to underpin confident human decision-making in this example:
\begin{itemize}
    \item \textit{traceability} in terms of salient features in the video and audio, for example, using the technique in~\cite{Hiley:2019} to discriminate the significant spatial and temporal features (in the latter case, the `violent' motions); 
    \item \textit{justification} of the inference as to the meaning of the activity (insight and foresight) assuming that the inference can be made by machine processing; and 
    \item \textit{assurance} that counterfactuals have been considered (possibilities of harmless vs aggressive action) potentially represented via the uncertainty-awareness methods from~\cite{Kaplan:2018}.
\end{itemize}

\section{Conclusion and Future Work}

In this paper, we have applied the coalition situational understanding concept to the problem of achieving layered ISR in multi-domain operations, specifically where artificial intelligence and machine learning services provide for improved distributed data analytics, and intelligence augmentation --- particularly the need for assured and explainable AI --- supports improved human-machine cognition. We focused on three elements in realising the layered ISR vision: human-machine teaming, dense urban terrain analysis, and enhanced asset interoperability, highlighting how the need for explainable AI in the multi-partner coalition setting is key.

Our current and future work focuses on the general problem illustrated in Figure~\ref{fig:nonhierarchical}: enabling subsymbolic AI agents to share uncertainty-aware representations of insights and knowledge that can then be communicated to symbolic AI agents, while also equipping symbolic AI agents to share insights to subsymbolic AI agents (i.e., machine-to-machine tellability). Ultimately, we seek to develop  techniques by which AI/ML agents can synergistically interact with humans via continuous learning from human-machine teaming activities.

\bibliographystyle{aaai}
\bibliography{why}

\end{document}